\newif\ifonecol
\onecolfalse%
\ifonecol
  \documentclass[10pt]{article}
  \usepackage[onecolumn]{btsstyle}
\else
  \documentclass[10pt,twocolumn]{article}
  \usepackage{btsstyle}
\fi

\newcommand{\pv}{p_{\mathrm{vis}}}
\newcommand{\pb}{p_{\mathrm{blind}}}
\newcommand{\lv}{\ell^{\mathrm{vis}}}
\newcommand{\lb}{\ell^{\mathrm{blind}}}
\newcommand{\BTS}{\mathrm{BTS}}
\newcommand{\JSD}{\mathrm{JSD}}
\newcommand{\pref}{\pi_{\mathrm{ref}}}
\newcommand{\ptheta}{\pi_{\theta}}
\graphicspath{{../figures/}{figures/}}
\usepackage{placeins}
\DeclareUnicodeCharacter{03BB}{\ensuremath{\lambda}}

\btstitle{BTS: Blind Trust Syndrome in Vision-Language Models (VLMs) via\\
          Pointwise Mutual Information and Image-Free Contrastive Decoding}
\btsauthor{%
  {\large\bfseries Minseok Shin}\\[4pt]
  Horcrux Technologies, Seoul, Republic of Korea\\
  \texttt{https://www.horcruxtechnologies.com}\\
  \texttt{horcruxtech.ai@gmail.com}\quad\texttt{stevenshin16@korea.ac.kr}}

\btsabstract{%
Vision-language models (VLMs) can answer as if they had not looked at the image. We study this failure, which we call \emph{Blind Trust Syndrome}, by comparing each next-token distribution with the one the same model produces when the image is removed. The resulting token-level \emph{BTS score} detects wrong answers on POPE with AUROC 0.71--0.77 on LLaVA-1.5-7B and rises along generated captions on content words, but a simpler pointwise mutual information (PMI) signal detects them significantly better (0.78--0.81). We show that contrastive decoding against the image-removed stream is exactly a PMI re-ranking, derive the condition under which it changes the output, and prove that a zero-threshold PMI gate preserves every answer flip when the image-free distribution is concentrated on the two answers. The experiments are consistent with this: across 7{,}200 POPE test questions the gated decoder, which acts only on content tokens with negative PMI, changes no yes/no answer relative to always-on correction, yet edits 70--76\% fewer responses. On LLaVA-1.5-7B, image-removed contrast raises adversarial POPE accuracy from 83.58\% to 84.50\% (exact McNemar $p=0.004$), whereas a noised-image contrast under the same greedy baseline lowers it to 80.71\%. Across LLaVA-1.5-7B, LLaVA-1.5-13B and LLaVA-NeXT, image-removed contrast is at least as accurate as greedy decoding on all nine model--split pairs and significantly better on four after Bonferroni correction, but the split that benefits differs by model. On LLaVA-1.5-7B the benefit does not transfer to long captions: neither form of contrast reduces CHAIR hallucination, and image-removed contrast significantly increases CHAIR\textsubscript{I}. Preference optimization against the model's own blind answers (PAPO) does not help: on LLaVA-1.5-7B it lowers POPE accuracy on two of three splits, does not significantly change CHAIR, and its token weighting has no measurable effect. In these experiments, image-removed contrast changed closed-form answers by small amounts and did not reduce caption hallucination, and the size of a reported gain can depend on the decoding baseline.}

\btskeywords{vision-language models; object hallucination; language priors; preference optimization; contrastive decoding; self-correction.}

\begin{document}
\btsmaketitle

\section{Introduction}
\label{sec:intro}

Large vision-language models (VLMs) such as LLaVA~\citep{liu2023llava,liu2024llava15} pair a vision encoder with a large language model and answer free-form questions about images. They can \emph{hallucinate}, mentioning objects, attributes or relations that the image does not contain~\citep{rohrbach2018chair,li2023pope,bai2024survey}. This limits their use in settings where the answer must be grounded in what is visible.

A recurring explanation is that the language model inside a VLM brings strong \emph{priors} that can override the visual input~\citep{agrawal2018dont,leng2024vcd}. If ``fork'' usually follows ``a plate with a knife and'', the model may write ``fork'' whether or not a fork is in the picture. We formalize this explanation as a measurable quantity. We call the failure mode \textbf{Blind Trust Syndrome (BTS)}: the model's next-token distribution with the image becomes almost the same as the distribution it would produce with the image removed, and the chosen token is one that the blind model already predicts with high confidence. In that regime the output depends little on the image.

Three questions remain open. \emph{Is blind trust a measurable predictor of error?} Decoding methods contrast the model with a perturbed or image-free copy of itself~\citep{leng2024vcd,wang2024icd,favero2024m3id,fang2025cpmi}, but they use the contrast as a correction and rarely validate it against ground truth at the level of individual answers. \emph{When does a contrastive rule actually change an output?} Existing work tunes its strength empirically; we give an exact criterion. \emph{Does the removal of the image help everywhere?} Reported gains are usually measured against sampled decoding and on selected settings, and \citet{yin2025mirage} attribute the POPE gains of VCD, ICD and SID to unidirectional shifts of the output distribution and to a plausibility constraint that turns sampling into greedy search. We compare image-removed and noised-image contrast, paired on the same questions, against a greedy baseline, and report where correction fails.

\begin{figure*}[t]
\centering
\resizebox{\textwidth}{!}{%
\begin{tikzpicture}[
  font=\small,
  box/.style={draw, rounded corners=3pt, align=center, minimum height=1.0cm, inner sep=4pt, line width=0.6pt},
  arr/.style={-{Stealth[length=2.2mm]}, line width=0.6pt}]
  \node[box, fill=gray!10, minimum width=2.4cm] (img) at (0, 1.2) {Image $I$};
  \node[box, fill=gray!10, minimum width=2.4cm] (q)   at (0,-1.2) {Question $x$,\\ prefix $y_{<t}$};
  \node[box, fill=blue!12,   minimum width=3.8cm] (vis) at (4.4, 1.2) {VLM \textbf{visual} stream\\ $\pv(\cdot \mid I, x, y_{<t})$};
  \node[box, fill=orange!18, minimum width=3.8cm] (bl)  at (4.4,-1.2) {VLM \textbf{blind} stream\\ $\pb(\cdot \mid x, y_{<t})$};
  \node[font=\scriptsize, text=black!60] at (4.4, 0) {same weights, image removed};
  \draw[arr] (img.east) -- (vis.west);
  \draw[arr] (q.east) -- (bl.west);
  \draw[arr] (q.north east) -- (vis.south west);
  \node[box, fill=red!10, minimum width=4.0cm] (bts) at (9.1, 0)
       {\textbf{Diagnose}: BTS (Eq.~\ref{eq:bts}) and PMI (Eq.~\ref{eq:pmi})\\ $\BTS_t = \pb(\hat y_t)\bigl(1-\JSD/\ln 2\bigr)$\\ $\mathrm{PMI}_t(\hat y_t)=\log\pv(\hat y_t)-\log\pb(\hat y_t)$};
  \draw[arr] (vis.east) -- (bts.north west);
  \draw[arr] (bl.east) -- (bts.south west);
  \node[diamond, draw, aspect=2.2, fill=yellow!20, inner sep=1.5pt, line width=0.6pt] (gate) at (13.9, 0) {gate open?};
  \draw[arr] (bts.east) -- (gate.west);
  \node[box, fill=green!12, minimum width=4.2cm] (tgsc) at (17.2, 1.2)
       {\textbf{Correct (decoding)}: TGSC\\ $(1+\alpha)\,\lv_t - \alpha\,\lb_t$};
  \node[box, fill=gray!8, minimum width=4.2cm] (keep) at (17.2,-1.2)
       {keep $\arg\max_v \pv(v)$\\ (no intervention)};
  \draw[arr] (gate.north) |- node[pos=0.3, left, font=\scriptsize] {yes} (tgsc.west);
  \draw[arr] (gate.south) |- node[pos=0.3, left, font=\scriptsize] {no} (keep.west);
  \node[box, minimum width=1.0cm] (out) at (20.7, 0) {$y_t$};
  \draw[arr] (tgsc.east) -- (out.north west);
  \draw[arr] (keep.east) -- (out.south west);
  \node[box, fill=violet!10, minimum width=21.6cm] (papo) at (10.0,-3.5)
       {\textbf{Correct (training)}: PAPO \quad preferred $y^{+}$ = reference answer \quad vs.\quad
        dispreferred $y^{-}$ = the model's own \emph{blind} response\\
        token weights $w_t = 1 + \lambda\,\BTS_t\,\mathbf{1}[y^{-}_t\in\mathcal{C}]$ on $y^{-}$ \quad $\Rightarrow$ \quad LoRA update of the VLM (Eq.~\ref{eq:papo})};
  \draw[arr, dashed] (bl.south) -- node[right, font=\scriptsize] {generates $y^{-}$} (bl.south |- papo.north);
  \draw[arr, dashed] (bts.south) -- node[right, font=\scriptsize] {weights $w_t$} (bts.south |- papo.north);
\end{tikzpicture}}
\caption{Overview. The same VLM is run with and without the image. The two streams give the BTS score and the PMI of the chosen token (\emph{diagnosis}). PMI or BTS gates a contrastive correction at decoding time (TGSC), and the BTS score weights a preference objective at training time (PAPO) (\emph{correction}).}
\label{fig:overview}
\end{figure*}
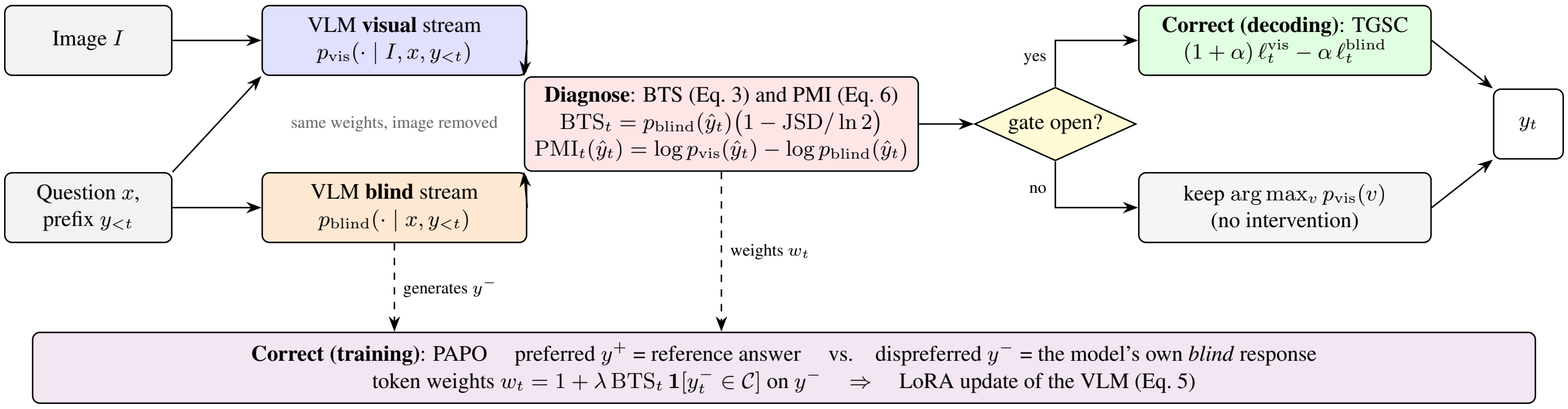

This paper examines image-removed contrast as a diagnostic, as a decoding rule and as a source of training negatives (Figure~\ref{fig:overview}). Our contributions are:
\begin{itemize}
  \setlength{\itemsep}{2pt}
  \item \textbf{A diagnostic tested against ground truth.} We define the BTS score and test it, together with PMI, against ground-truth POPE answers with bootstrap intervals and within each answer polarity (Sections~\ref{sec:diagnosis} and~\ref{sec:res-diag}). On LLaVA-1.5-7B, PMI has the higher AUROC on every split.
  \item \textbf{An exact flip criterion.} Image-removed contrast equals PMI re-ranking; we characterize when it changes a token and prove that a zero-threshold PMI gate is lossless when the image-free distribution is concentrated on two answers (Propositions~\ref{prop:flip} and~\ref{prop:binary}).
  \item \textbf{A controlled comparison.} On 7{,}200 POPE questions and 500 COCO captions from LLaVA-1.5-7B, and on POPE for two further models, we compare greedy decoding, noised-image contrast, always-on image-removed contrast and gated correction, paired per sample (Sections~\ref{sec:res-main} and~\ref{sec:res-chair}). The gate preserves all yes/no answers while editing far fewer responses, and neither contrast improves captions.
  \item \textbf{Blind-negative preference optimization (PAPO).} A DPO objective with the model's own blind answers as dispreferred samples and BTS-weighted rejected tokens (Section~\ref{sec:papo}). On LLaVA-1.5-7B it lowers POPE accuracy and does not significantly change CHAIR, and the BTS weighting has no measurable effect; we report this as a negative result.
\end{itemize}

\paragraph{Organization.} Section~\ref{sec:related} positions the work; Section~\ref{sec:background} fixes notation; Section~\ref{sec:method} defines the BTS score, PAPO and TGSC; Sections~\ref{sec:setup}--\ref{sec:analysis} give the setup, results and analysis; Section~\ref{sec:discussion} discusses limitations. Proofs and full tables are in the appendix.

\section{Related Work}
\label{sec:related}

\paragraph{Hallucination in VLMs.} Object hallucination was first quantified for image captioning with CHAIR~\citep{rohrbach2018chair}; POPE~\citep{li2023pope} turned it into a balanced yes/no probe with random, popular and adversarial negatives, and AMBER~\citep{wang2023amber}, HallusionBench~\citep{guan2024hallusionbench} and MMHal-Bench~\citep{sun2023llavarlhf} extend the evaluation to attributes, relations and visual illusions. Surveys~\citep{bai2024survey} identify over-reliance on language priors as a leading cause, echoing earlier findings in visual question answering~\citep{agrawal2018dont,goyal2017vqav2}.

\paragraph{Training-based mitigation.} LRV-Instruction~\citep{liu2024lrv} adds negative instructions; LLaVA-RLHF~\citep{sun2023llavarlhf} and RLHF-V~\citep{yu2024rlhfv} align with human or fact-augmented feedback; HA-DPO~\citep{zhao2023hadpo} and POVID~\citep{zhou2024povid} apply direct preference optimization (DPO)~\citep{rafailov2023dpo} with rewritten or AI-generated dispreferred answers, POVID also using distorted images. V-DPO~\citep{xie2024vdpo} adds a guidance term based on a zero-image distribution to the DPO objective at the sequence level. PGPO~\citep{ye2026pgpo} reshapes the policy-gradient advantage of each token by its KL visual dependency.

\paragraph{Decoding-based mitigation.} Contrastive decoding~\citep{li2023contrastive} subtracts a weaker distribution from a stronger one. VCD~\citep{leng2024vcd} contrasts the model with itself on a noised image and ICD~\citep{wang2024icd} on a disturbed instruction. M3ID~\citep{favero2024m3id} contrasts with the model without the image, applies the correction only when the conditioned model's top probability is below a threshold, increases the correction weight along the sequence, and also combines the rule with DPO. C-PMI~\citep{fang2025cpmi} derives the same logit form from conditional pointwise mutual information and adds a learned visual-token purifier. OPERA~\citep{huang2024opera} penalizes an attention pattern of ``over-trust'' in summary tokens during beam search, HALC~\citep{chen2024halc} contrasts fields of view with the help of a detector, and PAI~\citep{liu2024pai} amplifies attention on image tokens and subtracts the output for text-only input. SelfVal~\citep{liu2026selfval} measures the step-wise JSD between image-conditioned and text-only distributions, reports that it falls after the early part of a caption, and then validates generated objects with a language-prior-free prompt. Recent decoders choose when or how to apply contrast. MoD~\citep{chen2025mod} compares full and attended image tokens by JSD and switches between complementary and contrastive decoding; CICD~\citep{zhao2025cicd} contrasts with an unrelated image under a JSD gate; Octopus~\citep{suo2025octopus} learns a per-step choice among contrastive methods; CHASD~\citep{huang2026chasd} perturbs salient visual tokens only when the top probability is below a threshold; and SDPR~\citep{chen2026sdpr} builds a language-prior reference from early-layer features under an $L_1$ gate.

\paragraph{Analyses of contrastive decoding.} \citet{yin2025mirage} study VCD, ICD and SID on LLaVA-1.5 and Qwen-VL and attribute their POPE gains to two factors unrelated to hallucination: unidirectional adjustment of the output distribution, and the plausibility constraint, which reduces sampling to greedy search. They also report no improvement on CHAIR and mostly no-to-yes answer changes for VCD. Our comparison against greedy decoding shares this concern, and our noised-image results on CHAIR also show no improvement. That study does not examine image-removed contrast or a gate, which we add together with per-answer flip counts and an exact flip criterion. Concurrently, Zhao et al.\ contrast grounded generation with a vision-ablated path to obtain a token-level grounding score that flags reliance on language priors and drives a decoding-time penalty~\citep{zhao2026safe}; we additionally validate such a score against ground-truth answers, compare it with PMI, and characterize exactly when the resulting correction changes an output.

\paragraph{Post-hoc correction.} Woodpecker~\citep{yin2023woodpecker} verifies claims with expert models, LURE~\citep{zhou2024lure} trains a revisor, and Volcano~\citep{lee2024volcano} revises with self-feedback. They act after generation and typically require additional models or passes.

\paragraph{How this work differs.} Image-removed contrast, gating on a confidence signal and its combination with DPO are all present in M3ID~\citep{favero2024m3id}, image-removed contrast also appears in PAI~\citep{liu2024pai}, and we do not claim them as new. Table~\ref{tab:related} summarizes the landscape. Our work differs in the question it asks and the evidence it provides. (i) \emph{Validation of the signal:} to our knowledge, M3ID, C-PMI, V-DPO, PGPO, CHASD and SDPR use a contrast or image/no-image gap without reporting how well it discriminates erroneous answers; we report answer-level AUROC with intervals and within answer polarity, and find that the simpler PMI has a higher AUROC than our composite score. (ii) \emph{An exact account:} we are not aware of a statement of when contrast changes a token in the works above; Propositions~\ref{prop:flip} and~\ref{prop:binary} do, and the observed equivalence of gated and always-on correction on POPE is consistent with them. (iii) \emph{The gate condition:} M3ID gates on the visual stream's confidence, so its gate does not depend on the blind stream, while ours tests whether the image lowered the chosen token's probability. CHASD also gates on confidence, and MoD, CICD and SDPR gate on an unsigned distance between whole distributions, which is not specific to the chosen token. (iv) \emph{Baselines and negative results:} greedy baselines are also used elsewhere~\citep{yin2025mirage,huang2026chasd}; we additionally pair every comparison per question and report that the same rule that helps on POPE does not help on CHAIR. (v) \emph{Training:} PGPO up-weights tokens with \emph{high} visual dependency across sampled responses in a policy-gradient objective; PAPO up-weights the \emph{prior-driven} tokens of a rejected, model-generated blind answer in DPO. OPERA's ``over-trust'' concerns an attention pattern over summary tokens, whereas the quantity here is defined on output distributions.

\begin{table*}[t]
\caption{Comparison with existing hallucination-mitigation methods. \checkmark: yes; $\triangle$: partly (the visual input is perturbed or replaced rather than removed, so the language prior is not isolated).}
\label{tab:related}
\centering\small
\begin{adjustbox}{max width=\textwidth}
\begin{tabular}{@{}llcccp{4.6cm}ccc@{}}
\toprule
Method & Venue & Train & Decode & Post-hoc & Detection signal & Prior contrast & Gated & Extra model \\
\midrule
LRV-Instruction~\citep{liu2024lrv} & ICLR'24 & \checkmark & & & robust instruction data & & & \checkmark \\
LLaVA-RLHF~\citep{sun2023llavarlhf} & Findings ACL'24 & \checkmark & & & reward model & & & \checkmark \\
RLHF-V~\citep{yu2024rlhfv} & CVPR'24 & \checkmark & & & human segment-level correction & & & \\
HA-DPO~\citep{zhao2023hadpo} & arXiv'23 & \checkmark & & & rewritten preference pairs & & & \checkmark \\
POVID~\citep{zhou2024povid} & arXiv'24 & \checkmark & & & distorted image $\to$ dispreferred & $\triangle$ & & \checkmark \\
VCD~\citep{leng2024vcd} & CVPR'24 & & \checkmark & & distorted-image contrast & $\triangle$ & & \\
ICD~\citep{wang2024icd} & Findings ACL'24 & & \checkmark & & disturbed-instruction contrast & $\triangle$ & & \\
M3ID~\citep{favero2024m3id} & CVPR'24 & \checkmark & \checkmark & & image vs.\ masked-image information & \checkmark & \checkmark & \\
C-PMI~\citep{fang2025cpmi} & arXiv'25 & & \checkmark & & conditional PMI, token purifier & \checkmark & & \\
SelfVal~\citep{liu2026selfval} & arXiv'26 & & & \checkmark & language-prior-free object validation & & & \\
V-DPO~\citep{xie2024vdpo} & Findings EMNLP'24 & \checkmark & & & zero-image guidance in DPO & \checkmark & & \checkmark \\
PGPO~\citep{ye2026pgpo} & arXiv'26 & \checkmark & & & per-token KL visual dependency & \checkmark & & \\
OPERA~\citep{huang2024opera} & CVPR'24 & & \checkmark & & attention over-trust pattern & & \checkmark & \\
HALC~\citep{chen2024halc} & ICML'24 & & \checkmark & & field-of-view contrast & & & \checkmark \\
PAI~\citep{liu2024pai} & ECCV'24 & & \checkmark & & image-attention amplification, text-only contrast & \checkmark & & \\
Octopus~\citep{suo2025octopus} & CVPR'25 & \checkmark & \checkmark & & learned per-step choice among contrastive methods & $\triangle$ & \checkmark & \checkmark \\
MoD~\citep{chen2025mod} & Findings ACL'25 & & \checkmark & & JSD of full vs.\ attended image tokens & & \checkmark & \\
CICD~\citep{zhao2025cicd} & arXiv'25 & & \checkmark & & unrelated-image contrast, JSD gate & $\triangle$ & \checkmark & \\
CHASD~\citep{huang2026chasd} & arXiv'26 & & \checkmark & & local token perturbation, confidence gate & $\triangle$ & \checkmark & \\
SDPR~\citep{chen2026sdpr} & arXiv'26 & & \checkmark & & early-layer prior reference, $L_1$ gate & $\triangle$ & \checkmark & \\
Woodpecker~\citep{yin2023woodpecker} & SCIS'24 & & & \checkmark & expert models + LLM & & & \checkmark \\
LURE~\citep{zhou2024lure} & ICLR'24 & & & \checkmark & hallucination statistics & & & \checkmark \\
Volcano~\citep{lee2024volcano} & NAACL'24 & \checkmark & & \checkmark & self-feedback & & & \\
\midrule
\textbf{This work} & -- & \checkmark & \checkmark & & \textbf{BTS and PMI (image vs.\ blind)} & \checkmark & \checkmark & \\
\bottomrule
\end{tabular}
\end{adjustbox}
\end{table*}

\section{Background}
\label{sec:background}

\subsection{Autoregressive VLMs}
A VLM such as LLaVA-1.5~\citep{liu2024llava15} encodes an image $I$ into visual tokens, prepends them to the text prompt $x$, and generates an answer $y=(y_1,\dots,y_T)$ token by token from a vocabulary $\mathcal{V}$. At step $t$ the model produces logits $\lv_t\in\mathbb{R}^{|\mathcal{V}|}$ and the distribution
\begin{equation}
  \pv(\cdot) = \operatorname{softmax}(\lv_t), \qquad \lv_t = f_\theta(I, x, y_{<t}).
  \label{eq:pv}
\end{equation}
Table~\ref{tab:notation} in Appendix~\ref{app:details} lists all symbols.

\subsection{The blind distribution}
Because the language model inside a VLM is a complete language model, the same network can be run \emph{without} the image by simply removing the image tokens from the prompt. This gives the \emph{blind} logits and distribution
\begin{equation}
  \pb(\cdot) = \operatorname{softmax}(\lb_t), \qquad \lb_t = f_\theta(\varnothing, x, y_{<t}),
  \label{eq:pb}
\end{equation}
conditioned on exactly the same question and prefix. We use $\pb$ as an estimate of the model's language prior for the next token. Unlike a noised or masked image~\citep{leng2024vcd}, removing the image leaves no visual input; the same idea underlies M3ID~\citep{favero2024m3id}. Both streams share all weights and keep separate KV caches, so the blind stream costs one extra text-only forward pass per step.

\section{Diagnosing and Correcting Blind Trust}
\label{sec:method}

\subsection{The BTS score}
\label{sec:diagnosis}

Let $\hat y_t = \arg\max_v \pv(v)$ be the token the visual stream would emit. Blind trust has two ingredients: the image changes the prediction little, and the blind model already favours the token. We combine them as follows.

\begin{definition}[BTS score]
\label{def:bts}
\begin{equation}
  \BTS_t = \pb(\hat y_t)\cdot\Bigl(1 - \frac{\JSD(\pv\,\|\,\pb)}{\ln 2}\Bigr),
  \label{eq:bts}
\end{equation}
where $\JSD(p\|q)=\tfrac12\mathrm{KL}(p\|m)+\tfrac12\mathrm{KL}(q\|m)$, $m=\tfrac12(p+q)$, in nats~\citep{lin1991divergence}.
\end{definition}

\begin{proposition}[Range and limits]
\label{prop:bts}
$0 \le \BTS_t \le \pb(\hat y_t) \le 1$. If $\pb(\hat y_t)>0$, then $\BTS_t = \pb(\hat y_t)$ if and only if $\pv=\pb$, and $\BTS_t = 0$ whenever $\pv$ and $\pb$ have disjoint supports or the blind model assigns $\hat y_t$ zero probability.
\end{proposition}

The proof (Appendix~\ref{app:proof}) uses $0\le\JSD\le\ln 2$. Softmax outputs have full support, so the zero cases are limits that are approached but not attained. A high score therefore requires \emph{both} that the image did not move the distribution and that the chosen token is already likely without it; a token that is common in language but strongly supported by the image (low agreement) and a token that the image changed (low $\pb(\hat y_t)$) both receive low scores. The score needs no training, no labels and no extra model.

\subsection{Training-time correction: Prior-Aware Preference Optimization}
\label{sec:papo}

\paragraph{Dispreferred samples from the blind model.} For a training example $(I,x,y^{+})$ with a reference answer $y^{+}$, we let the model answer \emph{without} the image by greedy decoding of $\pb$, which gives $y^{-}$, and keep the pair unless $y^{-}$ equals or contains $y^{+}$ after VQA-style normalization. The blind response is generated without visual input, and no human annotation or external model is needed. Negatives from a model with masked visual tokens also appear in M3ID+DPO~\citep{favero2024m3id}; the addition here is the token-level weighting below.

\paragraph{BTS-weighted objective.} Let $\pref$ be the frozen initial model and $\ptheta$ the model being trained (LoRA~\citep{hu2022lora} on the language model; the reference is the same network with the adapter disabled). With $\log\pi(y\mid I,x)=\sum_t \log\pi(y_t\mid I,x,y_{<t})$ and per-token weights on the dispreferred response
\begin{equation}
  w_t = 1 + \lambda\,\BTS_t^{\mathrm{ref}}(y^{-}_t)\,\mathbf{1}[y^{-}_t\in\mathcal{C}], \qquad \lambda\ge 0,
  \label{eq:weights}
\end{equation}
where $\BTS_t^{\mathrm{ref}}(y^{-}_t)$ is Eq.~\eqref{eq:bts} with $\hat y_t$ replaced by $y^{-}_t$, computed with the reference model and both streams teacher-forced on $y^{-}_{<t}$, PAPO minimizes
\begin{equation}
\begin{split}
  \mathcal{L}_{\mathrm{PAPO}} = -\log\sigma\Bigl(\beta\Bigl[&\log\tfrac{\ptheta(y^{+}\mid I,x)}{\pref(y^{+}\mid I,x)}\\
   &- \textstyle\sum_t w_t \log\tfrac{\ptheta(y^{-}_t\mid I,x,y^{-}_{<t})}{\pref(y^{-}_t\mid I,x,y^{-}_{<t})}\Bigr]\Bigr).
\end{split}
\label{eq:papo}
\end{equation}
With $\lambda=0$, Eq.~\eqref{eq:papo} reduces to DPO~\citep{rafailov2023dpo} with blind negatives; $\lambda>0$ adds weight to the content tokens of $y^{-}$ in proportion to their BTS score. Both responses include the end-of-sequence token. In our training run the increase is small: the mean of $w_t$ over tokens and steps is 1.03 at $\lambda=1$, so the objective stays close to the case $\lambda=0$ (Figure~\ref{fig:train}).

\subsection{Decoding-time correction: Trust-Gated Self-Correction}
\label{sec:tgsc}

At inference time we run both streams at every step. Define the pointwise mutual information of a token,
\begin{equation}
  \mathrm{PMI}_t(v) = \log\pv(v) - \log\pb(v).
  \label{eq:pmi}
\end{equation}
A negative value means that seeing the image \emph{lowered} the token's probability. We use the term by analogy: a true conditional PMI has the marginal over images in its denominator, whereas $\pb$ is the model's output with no image, which only stands in for that marginal. A negative PMI is also a different condition from blind trust as defined above. Blind trust describes an image that changed the distribution little ($\pv\approx\pb$, so PMI is near zero), whereas $\mathrm{PMI}_t(\hat y_t)<0$ means that the image moved probability away from a token that is still ranked first.

\paragraph{Content tokens.} Sub-word continuations (``c\,|\,ouch'') and function words (``the'', ``on'') are largely fixed by the preceding text. We therefore restrict the gate to the set $\mathcal{C}$ of \emph{content tokens}: word-initial alphabetic tokens that are not function words, and digits; ``yes'' and ``no'' are retained.

\paragraph{Correction rule.} The gate is open when $\hat y_t\in\mathcal{C}$ and either $\BTS_t>\tau$ (BTS gate) or $\mathrm{PMI}_t(\hat y_t)<-\tau$ (PMI gate). If it is closed, the visual prediction is kept. Otherwise we apply a contrastive correction~\citep{li2023contrastive,leng2024vcd,favero2024m3id}, restricted to a plausible set:
\begin{equation}
\begin{gathered}
  \tilde\ell_t = (1+\alpha)\,\lv_t - \alpha\,\lb_t,\\
  \mathcal{V}_t = \{v : \pv(v) \ge \gamma \max_{u}\pv(u)\},\\
  y_t = \arg\max_{v\in\mathcal{V}_t} \tilde\ell_t(v).
\end{gathered}
\label{eq:tgsc}
\end{equation}
Algorithm~\ref{alg:tgsc} gives one decoding step. Both streams are evaluated at every step, so the gate does not reduce computation; it only decides whether the correction is used.

\paragraph{When does the correction change a token?} Up to an additive constant shared by all tokens, $\tilde\ell_t(v)=\log\pv(v)+\alpha\,\mathrm{PMI}_t(v)$, which has the form of classifier-free guidance~\citep{ho2022cfg}. This gives an exact criterion.

\begin{proposition}[Flip criterion]
\label{prop:flip}
Let $\alpha>0$, $\gamma\in(0,1]$, $\Delta_t(v)=\log\pv(\hat y_t)-\log\pv(v)\ge 0$, and let ties be broken in favour of $\hat y_t$. Applying Eq.~\eqref{eq:tgsc} changes the output token if and only if some $v\in\mathcal{V}_t\setminus\{\hat y_t\}$ satisfies
\begin{equation}
  \mathrm{PMI}_t(v)-\mathrm{PMI}_t(\hat y_t) > \Delta_t(v)/\alpha .
  \label{eq:flip}
\end{equation}
In particular, if $\pv=\pb$ the output never changes.
\end{proposition}

Membership in $\mathcal{V}_t$ also bounds the deficit: $\Delta_t(v)\le\ln(1/\gamma)$, about $2.30$ for $\gamma=0.1$. A token with no competitor in $\mathcal{V}_t$ therefore cannot change at any strength $\alpha$. A gate can only remove steps at which the correction applies. At a fixed prefix, every token changed by the gated rule is therefore also changed by the always-on rule, and the gate can improve on always-on correction only by blocking harmful changes. When the image-free distribution is concentrated on two answers, the correspondence is tighter.

\begin{proposition}[Two-way answers and residual mass]
\label{prop:binary}
For $v\neq\hat y_t$ let $R_t(v)=1-\pb(\hat y_t)-\pb(v)$. If the correction changes $\hat y_t$ to $v$ while $\mathrm{PMI}_t(\hat y_t)\ge 0$, then $\pv(v)>\pb(v)$ and
\begin{equation}
  \bigl[\pv(\hat y_t)-\pb(\hat y_t)\bigr]+\bigl[\pv(v)-\pb(v)\bigr]\le R_t(v).
  \label{eq:residual}
\end{equation}
In particular, if $\pb$ is supported on $\{\hat y_t,v\}$, so that $R_t(v)=0$, every such change implies $\mathrm{PMI}_t(\hat y_t)<0$. If also $\hat y_t\in\mathcal{C}$, the PMI gate with $\tau=0$ is open at every step where always-on correction would change the token.
\end{proposition}

Softmax outputs have full support, so $R_t(v)=0$ holds only approximately. By Eq.~\eqref{eq:residual}, the gate can miss a flip only when the blind model places enough probability outside the two answers. We did not log $R_t$, so this concentration is an assumption and not a measurement; at the answer token of POPE, which is the first generated token, the observed agreement of gated and always-on answers is consistent with it. Proofs are in Appendix~\ref{app:proof2}.

\begin{algorithm}[t]
\caption{One decoding step of TGSC}
\label{alg:tgsc}
\small
\KwIn{caches of the visual and blind streams; threshold $\tau$; strength $\alpha$; plausibility $\gamma$}
$\lv_t \leftarrow f_\theta(I,x,y_{<t})$;\quad $\lb_t \leftarrow f_\theta(\varnothing,x,y_{<t})$ \tcp*{two streams}
$\hat y_t \leftarrow \arg\max_v \pv(v)$\;
$s_t \leftarrow \BTS_t$ by Eq.~\eqref{eq:bts}, or $-\mathrm{PMI}_t(\hat y_t)$ by Eq.~\eqref{eq:pmi}\;
\eIf{$s_t > \tau$ \textbf{and} $\hat y_t\in\mathcal{C}$}{
  $y_t \leftarrow \arg\max_{v\in\mathcal{V}_t}\bigl[(1+\alpha)\lv_t(v)-\alpha\lb_t(v)\bigr]$ \tcp*{correct}
}{
  $y_t \leftarrow \hat y_t$ \tcp*{keep}
}
append $y_t$ to both streams\;
\end{algorithm}

\subsection{Metrics}
\label{sec:metrics}
\paragraph{Detection.} An answer is labelled as an error when it disagrees with the POPE ground truth. We report the AUROC of the BTS score and of $-\mathrm{PMI}$ at the answer token, also separately for ``yes'' and ``no'' answers, because answer polarity alone predicts errors (Table~\ref{tab:auroc}).
\paragraph{Hallucination.} POPE accuracy, F1, yes-ratio, false positives (FP, ``yes'' on an absent object) and false negatives (FN)~\citep{li2023pope}; CHAIR\textsubscript{S}, CHAIR\textsubscript{I} and object recall~\citep{rohrbach2018chair}.
\paragraph{Cost and side effects.} Gate rate (fraction of generated tokens at which the correction is applied, whether or not it changes the token), latency per answer, and the number of responses whose text differs from greedy decoding.

\section{Experimental Setup}
\label{sec:setup}

\paragraph{Models.} Three VLMs in bf16 without quantization: LLaVA-1.5-7B and LLaVA-1.5-13B~\citep{liu2024llava15} and LLaVA-NeXT-Mistral-7B (\texttt{llava-v1.6-mistral-7b}). LLaVA-1.5-7B is the main model. The other two are used for diagnosis and POPE only (Greedy, noised-image contrast, TGSC), with CHAIR, PAPO and test-set always-on correction not run.

\paragraph{Baselines.} Greedy decoding; a VCD-style noised-image contrast~\citep{leng2024vcd} in the same decoding loop (the released diffusion-noise schedule, noise step 500, $\alpha=1$, plausibility $0.1$, greedy selection); and always-on contrast against the blind stream, applied at every token including non-content tokens. The noise step differs from the 999 used for POPE in the VCD paper, and that paper reports sampling, so we refer to this baseline as \emph{noised-image contrast} and do not present it as a reproduction of VCD. The baseline was not tuned on validation data.

\paragraph{Training data.} 3{,}000 open-ended question--answer pairs from the VQAv2 validation split~\citep{goyal2017vqav2} (yes/no questions removed). Every question carries LLaVA-1.5's short-answer instruction, so preferred and dispreferred answers share one format. Blind responses are generated greedily; refusals, non-English and over-long outputs are discarded, and pairs whose blind answer matches or contains the reference after VQA-style normalization are dropped. The 3{,}000 questions yield 2{,}347 preference pairs; among blind answers that pass the validity filters, 21.5\% already match the reference and are dropped. VQAv2 validation images come from COCO val2014, as do the POPE and CHAIR images, and we did not remove images shared with the evaluation sets.

\paragraph{PAPO.} LoRA ($r=16$, $\alpha_{\mathrm{LoRA}}=32$, dropout $0.05$) on the query, key, value and output projections of the language model; AdamW, learning rate $5\times10^{-5}$, $\beta=0.1$, $\lambda\in\{1,0\}$, gradient accumulation 8, one epoch, one run per $\lambda$.

\paragraph{Decoding selection.} The gate signal and threshold (BTS: $\tau\in\{0.2,0.35,0.5,0.65,0.8\}$; PMI: $\tau\in\{0,0.1,0.25,0.5\}$) are selected first at $\alpha=1$, $\gamma=0.1$, then $\alpha\in\{1,2,3\}$ and $\gamma\in\{0.1,0.01\}$, by F1 on 600 POPE-adversarial \emph{validation} questions disjoint from the test questions (300 questions and only $\alpha=1$, $\gamma=0.1$ on NeXT); ties are broken by lower harm, higher help, then lower gate rate. The selected setting on LLaVA-1.5-7B is the PMI gate with $\tau=0$, $\alpha=3$, $\gamma=0.1$; on 13B and NeXT it is the PMI gate with $\tau=0$, $\alpha=1$, $\gamma=0.1$. The always-on baseline uses the same $\alpha,\gamma$.

\paragraph{Evaluation.} POPE~\citep{li2023pope}: of the 3{,}000 questions per split, 2{,}400 form the test set on LLaVA-1.5-7B/13B (1{,}000 on NeXT); prompt suffix ``Please answer yes or no.'' Diagnosis uses the first 1{,}000 test questions per split on LLaVA-1.5-7B (300 on 13B and NeXT) under greedy decoding and involves no selection. CHAIR~\citep{rohrbach2018chair}: 500 random COCO val2014 images, ``Please describe this image in detail.'', at most 512 new tokens; ground-truth objects are the union of instance annotations and objects mentioned in reference captions, using our re-implementation of the synonym mapping, so absolute CHAIR values are not comparable with other papers.

\paragraph{Statistics.} All comparisons are paired on the same questions or images. We report 2{,}000-resample bootstrap intervals for AUROC and for the CHAIR\textsubscript{I} difference, and exact two-sided McNemar tests for POPE accuracy and CHAIR\textsubscript{S}. POPE splits share their positive questions, so we test each split separately and do not pool them; Bonferroni-corrected values across the three splits are given where we claim significance.

\paragraph{Hardware and software.} One NVIDIA A100 (Google Colab), PyTorch and Hugging Face \texttt{transformers}; every number is computed from logged per-sample results.

\section{Results}
\label{sec:results}

\subsection{Detecting wrong answers from the image/no-image gap}
\label{sec:res-diag}

\begin{figure*}[t]
\centering
\includegraphics[width=\textwidth]{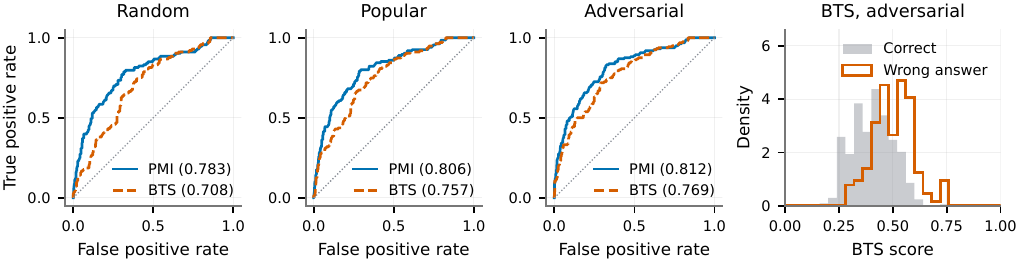}
\caption{Detection of wrong POPE answers on LLaVA-1.5-7B (1{,}000 questions per split). Left three panels: ROC of $-$PMI and BTS at the answer token, with AUROC in the legend. Right: BTS at the answer token for correct and wrong answers on the adversarial split.}
\label{fig:diag}
\end{figure*}

Table~\ref{tab:auroc} and Figure~\ref{fig:diag} show that both signals carry information about answer errors. BTS reaches an AUROC of 0.71--0.77. Answering ``no'' alone predicts errors with AUROC 0.56--0.67, so we also report each signal within one answer polarity, where it cannot exploit that shortcut. PMI is significantly better than BTS on all splits (paired bootstrap intervals of the difference $[0.052,0.104]$, $[0.030,0.070]$ and $[0.023,0.063]$ for random, popular and adversarial), and the two signals are highly correlated (Pearson $r\approx 0.91$ between BTS and $-$PMI). On these data, the product form of BTS therefore adds no detection power over the simpler log-ratio. Appendix Figure~\ref{fig:token} shows the score on one caption.

\begin{table*}[t]
\caption{Detection of erroneous POPE answers at the answer token (LLaVA-1.5-7B, greedy; 1{,}000 questions per split). Brackets: 95\% bootstrap interval. ``Ans.\ no'': AUROC of the answer polarity alone.}
\label{tab:auroc}
\centering\small
\begin{tabular}{@{}lccccccc@{}}
\toprule
Split & Errors (FP / FN) & BTS & PMI & BTS ``yes'' & PMI ``yes'' & BTS ``no'' & PMI ``no'' \\
\midrule
Random & 113 (14 / 99) & 0.708 [0.662, 0.755] & 0.783 [0.736, 0.830] & 0.865 & 0.912 & 0.706 & 0.783 \\
Popular & 135 (36 / 99) & 0.757 [0.716, 0.796] & 0.806 [0.766, 0.845] & 0.855 & 0.871 & 0.760 & 0.804 \\
Adversarial & 160 (59 / 101) & 0.769 [0.731, 0.808] & 0.812 [0.775, 0.848] & 0.859 & 0.892 & 0.717 & 0.758 \\
\bottomrule
\multicolumn{8}{@{}l}{Ans.\ no: 0.666 (random), 0.600 (popular), 0.555 (adversarial).}\\
\end{tabular}
\end{table*}

\begin{figure}[t]
\centering
\includegraphics[width=\linewidth]{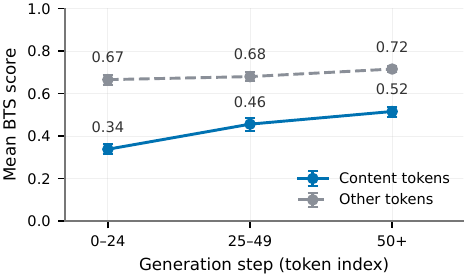}
\caption{Mean BTS by position in 50 captions, pooled over tokens. Bars: 95\% bootstrap interval, resampling captions.}
\label{fig:captionbts}
\end{figure}

\paragraph{BTS along a caption.} On 50 captions (Figure~\ref{fig:captionbts}) of at most 96 tokens, mean BTS on content tokens rises from 0.34 (first 25 tokens) to 0.46 (tokens 25--49) and 0.52 (tokens 50 onward); Spearman $\rho=0.24$ over 1{,}783 content tokens, and the per-caption slope is positive for 94\% of captions. Non-content tokens are high from the start and change little (0.67, 0.68, 0.72; $\rho=0.06$). The trend agrees with the fall in JSD reported by~\citet{liu2026selfval} with the increasing weight used by M3ID~\citep{favero2024m3id}, and with the observation in LURE~\citep{zhou2024lure} that hallucination often appears late in the text. It is a correlation with position in 50 captions, and does not by itself show that late tokens are more often hallucinated.

\subsection{POPE}
\label{sec:res-main}

Tables~\ref{tab:main} and~\ref{tab:paired} report POPE on LLaVA-1.5-7B. Accuracy under image-removed contrast is higher on all three splits, by 0.25, 0.63 and 0.92 points. The improvement on the adversarial split survives Bonferroni correction ($p=0.012$); the popular split does not ($p=0.095$), and the random split is not significant. Most remaining errors are false negatives (about 250 per split), which the correction hardly changes; the gain comes mainly from false positives on the popular and adversarial splits (78 to 67 and 141 to 125).

\begin{table*}[t]
\caption{POPE on LLaVA-1.5-7B (2{,}400 test questions per split). TGSC: PMI gate, $\tau=0$, $\alpha=3$, $\gamma=0.1$. Latency: per answer.}
\label{tab:main}
\centering\small
\begin{tabular}{@{}l*{12}{c}@{}}
\toprule
& \multicolumn{4}{c}{Random} & \multicolumn{4}{c}{Popular} & \multicolumn{4}{c}{Adversarial} \\
Method & Acc. & F1 & FP & FN & Acc. & F1 & FP & FN & Acc. & F1 & FP & FN \\
\midrule
Greedy & 88.17 & 86.86 & 33 & 251 & 86.29 & 85.09 & 78 & 251 & 83.58 & 82.63 & 141 & 253 \\
Noised-image contrast & 87.50 & 86.39 & 62 & 238 & 84.54 & 83.66 & 131 & 240 & 80.71 & 80.36 & 220 & 243 \\
Image-removed, always-on & 88.42 & 87.17 & 32 & 246 & 86.92 & 85.73 & 67 & 247 & 84.50 & 83.53 & 125 & 247 \\
TGSC (PMI gate) & 88.42 & 87.17 & 32 & 246 & 86.92 & 85.73 & 67 & 247 & 84.50 & 83.53 & 125 & 247 \\
\bottomrule
\end{tabular}
\end{table*}

\begin{table*}[t]
\caption{Paired comparisons on POPE (LLaVA-1.5-7B). ``Improved / worsened'': questions correct only under the method / only under the baseline. $p$: exact McNemar, not corrected for multiplicity. ``Edited'': responses whose text differs from greedy.}
\label{tab:paired}
\centering\small
\begin{tabular}{@{}lccccc@{}}
\toprule
Comparison & Split & Improved / worsened & $p$ & Edited (always $\to$ TGSC) & no$\to$yes / yes$\to$no \\
\midrule
Image-removed vs.\ greedy & Random & 17 / 11 & 0.345 & 331 $\to$ 81 & 16 / 12 \\
 & Popular & 29 / 14 & 0.032 & 328 $\to$ 97 & 18 / 25 \\
 & Adversarial & 38 / 16 & 0.0038 & 399 $\to$ 121 & 22 / 32 \\
\midrule
Noised-image vs.\ greedy & Random & 46 / 62 & 0.149 & -- & 75 / 33 \\
 & Popular & 48 / 90 & $4.4\times10^{-4}$ & -- & 101 / 37 \\
 & Adversarial & 52 / 121 & $1.6\times10^{-7}$ & -- & 131 / 42 \\
\midrule
Image-removed vs.\ noised-image & Random & 71 / 49 & 0.055 & -- & -- \\
 & Popular & 106 / 49 & $5.4\times10^{-6}$ & -- & -- \\
 & Adversarial & 140 / 49 & $2.4\times10^{-11}$ & -- & -- \\
\bottomrule
\end{tabular}
\end{table*}

\begin{figure}[t]
\centering
\includegraphics[width=\linewidth]{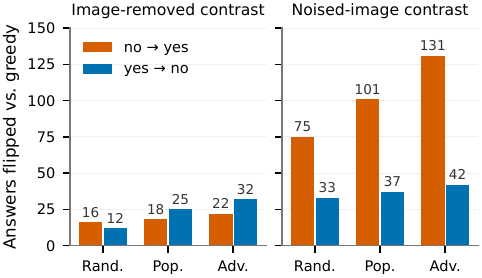}
\caption{Answers changed relative to greedy decoding, by direction (LLaVA-1.5-7B, 2{,}400 questions per split).}
\label{fig:flips}
\end{figure}

\paragraph{The gate preserves answers and removes edits.} TGSC and always-on correction give the same yes/no answer on all 7{,}200 questions, which is consistent with Proposition~\ref{prop:binary}. They are not the same decoder: TGSC edits 70--76\% fewer responses, with a gate rate of 3.4--4.3\% of generated tokens. Always-on correction applies at every token, whereas TGSC applies only at content tokens with negative PMI, so this reduction reflects both conditions. The extra edits of always-on correction do not alter the answer, so on POPE they bring no measurable benefit.

\paragraph{Noised-image contrast.} Under our configuration, noised-image contrast is significantly worse than greedy decoding on the popular and adversarial splits. It flips answers mostly toward ``yes'' (no$\to$yes outnumbers yes$\to$no 3:1 on the adversarial split; Figure~\ref{fig:flips}) and raises false positives from 141 to 220, while image-removed flips are balanced. The two contrasts are also largely disjoint: only 19 of the 54 adversarial flips of image-removed contrast are also flipped by noised-image contrast. Our noise step (500), strength ($\alpha=1$) and greedy selection differ from the POPE setting reported for VCD, so these results describe this configuration and not VCD itself. We note that the VCD paper reports sampled decoding as its baseline (83.29/81.88/78.96\% on random/popular/adversarial) and 87.73/85.38/80.88\% with VCD. Our greedy baseline already reaches 88.17/86.29/83.58\%. The question sets and implementations differ, so this is not a direct comparison, but it shows that the choice of baseline can account for a difference of this size, in line with~\citet{yin2025mirage}.

\paragraph{Cost.} Latency per answer is 85\,ms for greedy, 110\,ms for image-removed contrast (+29\%), and 156\,ms for noised-image contrast (+84\%), which needs a second image-conditioned pass. TGSC and always-on correction cost the same, since both run the blind stream at every step.

\paragraph{Generalization across models.} Tables~\ref{tab:models} and~\ref{tab:auroc2} repeat the comparison on LLaVA-1.5-13B and LLaVA-NeXT-Mistral-7B. PMI has a higher AUROC than BTS on all six model--split pairs, and the interval of the difference excludes zero on five of them; with 300 questions per split these estimates are coarse. Image-removed contrast is never below greedy decoding. After Bonferroni correction across the three splits of each model, it is significantly better on the popular ($p=0.039$) and adversarial ($p=0.018$) splits of 13B and on the random split of NeXT ($p=0.035$). The split that benefits most differs by model: adversarial on 7B and 13B, and random on NeXT, whose adversarial gain is 0.10 points ($p=1.0$). The type of error also differs. On NeXT, all 69 answer flips are no$\to$yes, so false negatives fall from 95 to about 80 while false positives rise; on 13B, flips go in both directions. Noised-image contrast is significantly worse than greedy decoding on all three splits of 13B and not significantly different on NeXT. On validation data, the zero-threshold PMI gate has the same metrics as always-on correction with a gate rate of 2.7\% (13B) and 5.6\% (NeXT); we did not run always-on correction on the test sets of these two models, so the answer-identity result of LLaVA-1.5-7B is not repeated there.

\begin{table*}[t]
\caption{POPE on LLaVA-1.5-13B (2{,}400 questions per split) and LLaVA-NeXT-Mistral-7B (1{,}000). Accuracy in \%. ``Impr.\ / wors.'': questions correct only under the method / only under greedy decoding. $p$: exact McNemar, not corrected for multiplicity. FP and FN: greedy $\to$ TGSC.}
\label{tab:models}
\centering\small
\begin{adjustbox}{max width=\textwidth}
\begin{tabular}{@{}llccccccccc@{}}
\toprule
Model & Split & Greedy & Noised & TGSC & TGSC impr.\ / wors. & $p$ & Noised impr.\ / wors. & $p$ & FP & FN \\
\midrule
13B & Random & 90.58 & 88.42 & 90.62 & 8 / 7 & 1.0 & 24 / 76 & $1.8\times10^{-7}$ & 59 $\to$ 63 & 167 $\to$ 162 \\
 & Popular & 87.88 & 85.46 & 88.29 & 12 / 2 & 0.013 & 31 / 89 & $1.1\times10^{-7}$ & 123 $\to$ 119 & 168 $\to$ 162 \\
 & Adversarial & 83.75 & 81.79 & 84.38 & 21 / 6 & 0.0059 & 49 / 96 & $1.2\times10^{-4}$ & 221 $\to$ 214 & 169 $\to$ 161 \\
\midrule
NeXT & Random & 89.90 & 90.10 & 91.10 & 16 / 4 & 0.012 & 14 / 12 & 0.85 & 6 $\to$ 10 & 95 $\to$ 79 \\
 & Popular & 89.10 & 88.00 & 89.90 & 15 / 7 & 0.13 & 15 / 26 & 0.12 & 14 $\to$ 21 & 95 $\to$ 80 \\
 & Adversarial & 86.50 & 85.60 & 86.60 & 14 / 13 & 1.0 & 21 / 30 & 0.26 & 40 $\to$ 53 & 95 $\to$ 81 \\
\bottomrule
\end{tabular}
\end{adjustbox}
\end{table*}

\begin{table}[t]
\caption{Detection of wrong POPE answers on the other two models (300 questions per split). Interval: paired 95\% bootstrap of PMI $-$ BTS.}
\label{tab:auroc2}
\centering\small
\begin{adjustbox}{max width=\linewidth}
\begin{tabular}{@{}llcccc@{}}
\toprule
Model & Split & Errors & BTS & PMI & Difference \\
\midrule
13B & Random & 22 & 0.723 & 0.850 & $[0.057, 0.207]$ \\
 & Popular & 27 & 0.818 & 0.879 & $[0.006, 0.123]$ \\
 & Adversarial & 44 & 0.834 & 0.870 & $[-0.005, 0.083]$ \\
\midrule
NeXT & Random & 25 & 0.862 & 0.944 & $[0.037, 0.134]$ \\
 & Popular & 30 & 0.760 & 0.835 & $[0.039, 0.118]$ \\
 & Adversarial & 36 & 0.711 & 0.766 & $[0.021, 0.094]$ \\
\bottomrule
\end{tabular}
\end{adjustbox}
\end{table}

\subsection{Free-form captioning (CHAIR)}
\label{sec:res-chair}

Table~\ref{tab:chair} reports CHAIR. No contrastive method reduces hallucination. On the base model, CHAIR\textsubscript{S} is higher than greedy for all contrastive methods, and none of those differences is significant. CHAIR\textsubscript{I} increases significantly under image-removed contrast, both always-on (difference interval $[0.003,0.047]$) and gated ($[0.001,0.041]$), and not under noised-image contrast ($[-0.015,0.023]$). Always-on correction also shortens captions by 16\% (83.6 to 69.8 words) while making CHAIR\textsubscript{I} worse. The gate restores the caption length (81.3 words), gives a lower CHAIR\textsubscript{S} than always-on correction, and gives the highest object recall (0.802 versus 0.791 for greedy), but it does not bring CHAIR\textsubscript{I} back to the greedy level.

\begin{table*}[t]
\caption{CHAIR on 500 COCO val2014 images (LLaVA-1.5-7B; lower CHAIR and higher recall are better). Fixed / new: captions that became hallucination-free / newly contain a hallucination, relative to greedy; $p$: exact McNemar on CHAIR\textsubscript{S}.}
\label{tab:chair}
\centering\small
\begin{adjustbox}{max width=\textwidth}
\begin{tabular}{@{}lccccccc@{}}
\toprule
Method & CHAIR\textsubscript{S} & CHAIR\textsubscript{I} & Recall & Words & Latency (ms) & Fixed / new & $p$ \\
\midrule
Greedy & 0.514 & 0.257 & 0.791 & 83.6 & 720 & -- & -- \\
Noised-image contrast & 0.526 & 0.260 & 0.784 & 82.6 & 1450 & 54 / 60 & 0.64 \\
Image-removed, always-on & 0.548 & 0.282 & 0.788 & 69.8 & 1226 & 66 / 83 & 0.19 \\
TGSC (PMI gate; gate rate 18.5\%) & 0.528 & 0.277 & 0.802 & 81.3 & 1362 & 61 / 68 & 0.60 \\
\midrule
PAPO ($\lambda=1$), greedy & 0.496 & 0.254 & 0.795 & 89.8 & 1608 & 66 / 57 & 0.47 \\
PAPO ($\lambda=1$) + TGSC (gate rate 15.8\%) & 0.540 & 0.283 & 0.805 & 88.6 & 2871 & 61 / 74 & 0.30 \\
\bottomrule
\end{tabular}
\end{adjustbox}
\end{table*}

\subsection{Training and decoding}
\label{sec:res-ablation}

\begin{figure}[t]
\centering
\includegraphics[width=\linewidth]{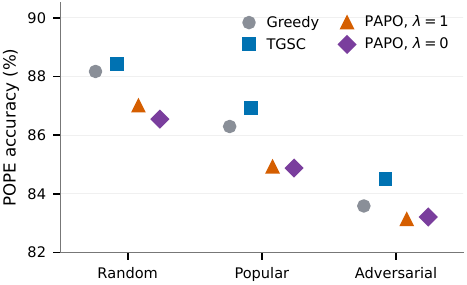}
\caption{POPE accuracy of the base model and of the PAPO models (LLaVA-1.5-7B). The vertical axis starts at 82\%. Adding TGSC to either PAPO model leaves accuracy unchanged, so those points are omitted.}
\label{fig:ablation}
\end{figure}

Figure~\ref{fig:ablation} and Table~\ref{tab:ablation} report PAPO on LLaVA-1.5-7B. PAPO lowers POPE accuracy relative to the base model: by 1.17, 1.37 and 0.45 points at $\lambda=1$, and by 1.63, 1.41 and 0.37 points at $\lambda=0$, significantly on the random and popular splits. The loss comes from false negatives, which rise from about 250 to 276 ($\lambda=1$) and 293--294 ($\lambda=0$); false positives fall only at $\lambda=0$ and on the adversarial split at $\lambda=1$. The two weightings do not differ significantly on any split ($p=0.11$, $1.0$, $0.89$), which agrees with the mean weight of 1.03: the BTS weighting has no measurable effect. On CHAIR (Table~\ref{tab:chair}), PAPO gives lower CHAIR\textsubscript{S} (0.496 versus 0.514) and longer captions, but neither CHAIR measure differs significantly from the base model ($p=0.47$; CHAIR\textsubscript{I} difference interval $[-0.023, 0.017]$). Adding TGSC to the PAPO model leaves every POPE answer unchanged, with a gate rate below 0.01\%. On CHAIR the gate still opens, and CHAIR\textsubscript{I} is significantly higher than for base greedy decoding ($[0.005, 0.045]$). We therefore find no complementarity between the two stages.

\begin{table*}[t]
\caption{Effect of PAPO on POPE (LLaVA-1.5-7B, 2{,}400 questions per split, greedy decoding). $p$: exact McNemar against the base model, not corrected for multiplicity.}
\label{tab:ablation}
\centering\small
\begin{tabular}{@{}l*{12}{c}@{}}
\toprule
& \multicolumn{4}{c}{Random} & \multicolumn{4}{c}{Popular} & \multicolumn{4}{c}{Adversarial} \\
Model & Acc. & FP & FN & $p$ & Acc. & FP & FN & $p$ & Acc. & FP & FN & $p$ \\
\midrule
Base & 88.17 & 33 & 251 & -- & 86.29 & 78 & 251 & -- & 83.58 & 141 & 253 & -- \\
PAPO, $\lambda=1$ & 87.00 & 36 & 276 & 0.0026 & 84.92 & 86 & 276 & 0.0017 & 83.13 & 129 & 276 & 0.39 \\
PAPO, $\lambda=0$ & 86.54 & 30 & 293 & $9.8\times10^{-6}$ & 84.88 & 69 & 294 & $9.9\times10^{-4}$ & 83.21 & 110 & 293 & 0.47 \\
\bottomrule
\end{tabular}
\end{table*}

\subsection{Sensitivity of the gate}
\label{sec:res-gate}

Appendix Table~\ref{tab:tau} sweeps the gate on the 600 validation questions (Appendix Figure~\ref{fig:tau}). Differences between settings are within one to three questions: F1 ranges from 0.811 to 0.814, and $\alpha=3$ is selected over $\alpha=1$ by a one-question difference (F1 0.8144 versus 0.8138). The selected values should therefore be read as one acceptable setting and not as an optimum. The BTS gate at $\tau\le 0.5$ and the PMI gate at $\tau=0$ give identical validation metrics, but the PMI gate is open far less often (gate rate 3.4\% versus 10.6--44.1\%). Raising either threshold removes the helpful flips first.

\subsection{Qualitative results}
\begin{figure*}[!t]
\centering
\includegraphics[width=\textwidth]{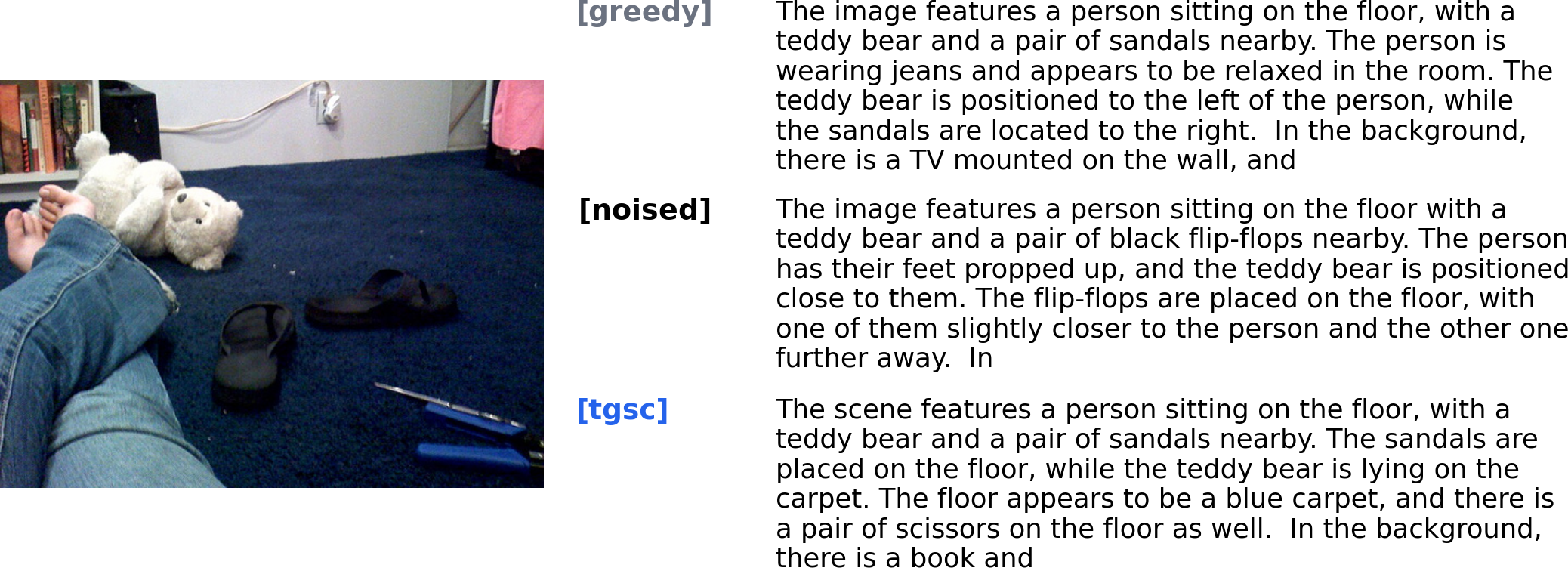}\\[6pt]
\includegraphics[width=\textwidth]{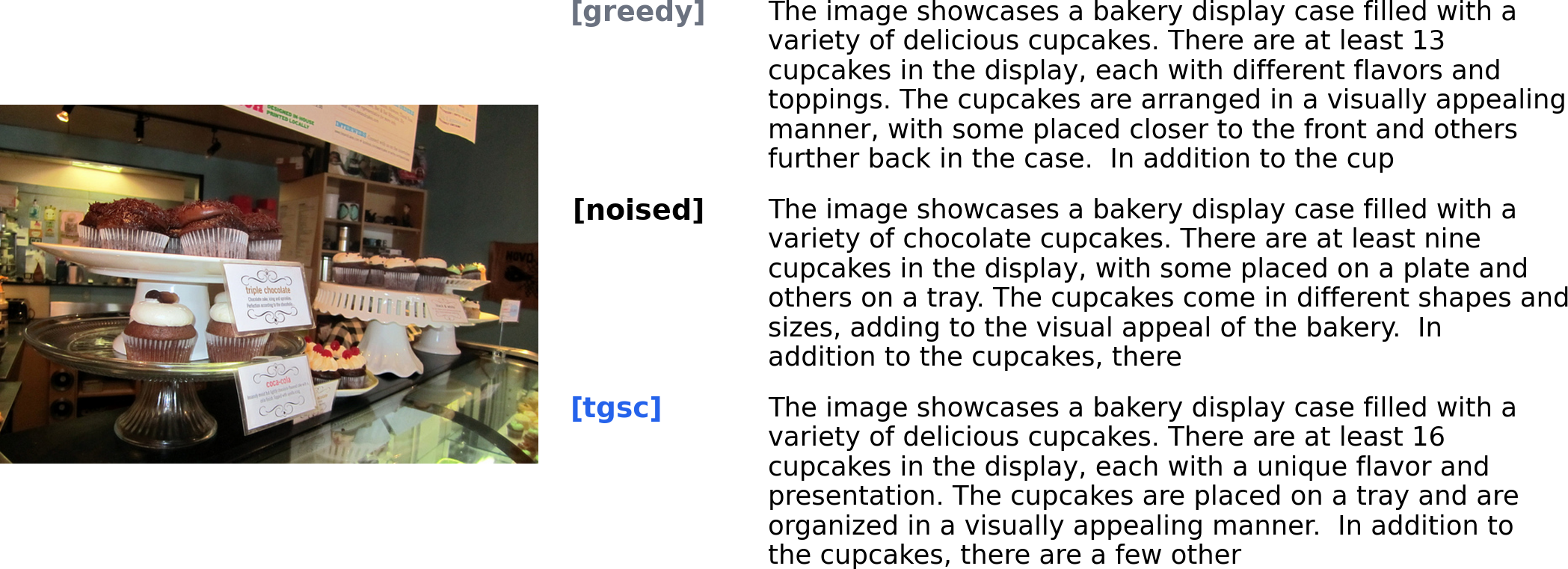}
\caption{Illustrative responses to ``Describe this image in detail.'', not representative samples; each stops at the 80-token limit. ``noised'' is the noised-image contrast. Top: greedy decoding mentions a TV that is not visible, and TGSC does not. Bottom: the decoders give different lower bounds on the number of cupcakes (13, nine and 16), which we did not verify.}
\label{fig:qual}
\end{figure*}

Figure~\ref{fig:qual} shows two of the three qualitative images generated by the pipeline; they illustrate the output and are not evidence of an effect. For examples that do not depend on our judgement, we take the lowest COCO image id in each CHAIR outcome. In the first image where TGSC adds a hallucinated object relative to greedy decoding (id 974), TGSC writes ``there are two bottles visible in the scene''. In the first image where it removes one (id 294), greedy decoding writes ``a sink is located towards the right side of the kitchen'' and TGSC does not mention a sink. Over the 500 images these outcomes occur 68 and 61 times (Table~\ref{tab:chair}).

\section{Insights}
\label{sec:analysis}
\paragraph{Contrast is a re-ranking of what the model already considers.} The criterion in Eq.~\eqref{eq:flip} explains why the effect is small: a token changes only if a competitor inside $\mathcal{V}_t$ gains more PMI than its log-probability deficit scaled by $1/\alpha$. An answer with no competitor in $\mathcal{V}_t$ cannot change at any strength. This may contribute to the roughly 250 false negatives per split on LLaVA-1.5-7B, which stay almost unchanged, although on NeXT they fall from 95 to about 80; we did not log competitor probabilities, so it remains a hypothesis.
\paragraph{Which distribution is subtracted matters.} Image-removed contrast and noised-image contrast change mostly different questions and shift the answers in different directions. The noised-image stream still contains an image-shaped input, and on LLaVA-1.5-7B and NeXT subtracting it pushes answers toward ``yes'', while on 13B the two directions are nearly even. This suggests that noised-image contrast is not simply an approximation of the prior.
\paragraph{Effect of the gate on LLaVA-1.5-7B.} In the POPE runs of LLaVA-1.5-7B the zero-threshold gate had no effect on the yes/no decision, consistent with Proposition~\ref{prop:binary}. Its value is in avoiding edits to the rest of the response.
\paragraph{The benefit depends on the model.} The split with the largest gain and the direction of the flips both change between models (Table~\ref{tab:models}). A result on one model and one split should therefore not be read as a general property of contrastive decoding. On NeXT all flips go from ``no'' to ``yes'' and false positives rise, which is the kind of unidirectional shift that \citet{yin2025mirage} warn can raise POPE accuracy without reducing hallucination, so the NeXT gain should not be read as fewer hallucinations. On LLaVA-1.5-7B flips go in both directions and false positives fall.
\paragraph{Why correction does not transfer to captions.} In a caption, the blind stream is conditioned on a long prefix that was itself written from the image, so it is no longer a pure language prior; a token that is well supported by that prefix is penalized even when correct. We hypothesize that this, rather than a lack of strength, explains the increase in CHAIR\textsubscript{I} and the shorter captions under always-on correction. We did not test this mechanism directly; M3ID's increasing weight points in the opposite direction, and it reports caption gains against a sampled baseline, so the role of baseline and weighting needs separate study.
\paragraph{What does PAPO change?} At the POPE answer token, PAPO raises the median PMI of the chosen token from about 0.6 to about 12.3 and lowers the median BTS from 0.41 to about $10^{-9}$, and almost no answer has negative PMI; $\lambda=0$ gives nearly the same values. The adapter therefore separates the two streams at the answer token, which is why the gate stays closed on POPE and TGSC adds nothing there. In this configuration, the gate therefore no longer acts at the POPE answer token. The preference margin keeps growing until the end of the epoch (Appendix Figure~\ref{fig:train}); we did not evaluate during training, so whether a shorter or weaker training run would behave differently remains open.

\section{Discussion and Limitations}
\label{sec:discussion}
\paragraph{Effect size.} The gain on POPE is at most 1.2 accuracy points across the three models and is significant on only some splits, and on LLaVA-1.5-7B it does not extend to CHAIR. On POPE, image-removed contrast did not lower accuracy on any split of the three models, at a latency increase of 17--30\%; on NeXT its gain comes with more false positives. These results do not show that decoding-time correction removes hallucination.
\paragraph{Cost.} The blind stream runs at every step, gated or not. A lazy evaluation strategy needs a gate that does not depend on the blind stream, as in M3ID.
\paragraph{Limitations.} (i) CHAIR and PAPO are run only on LLaVA-1.5-7B. The other two models have POPE only, smaller diagnostic samples (300 questions per split; NeXT POPE uses 1{,}000 per split) and no test-set always-on run, and NeXT was tuned only at $\alpha=1$, $\gamma=0.1$. (ii) POPE probes object existence; AMBER, HallusionBench and general-capability benchmarks (MME, MMBench) were not run. (iii) The noised-image baseline is not tuned and does not match the VCD paper's POPE configuration; OPERA and other methods were not re-implemented. (iv) CHAIR uses our re-implementation and 500 images. (v) FP counts on the random split are small (14 in the diagnostic subset), so polarity-specific AUROCs there are uncertain. (vi) The blind distribution captures the prior given the text; priors induced by visible objects are not isolated. (vii) The training set is small and drawn from VQAv2. (viii) PAPO was trained once per $\lambda$ with one epoch and no early stopping, so the negative result covers this configuration only.

\section{Conclusion}
\label{sec:conclusion}
We measured how much a VLM's next-token distribution depends on the image and tested three uses of that measurement. As a detector, the image/no-image gap identifies wrong answers, with PMI obtaining a higher AUROC than our composite BTS score on all nine model--split pairs. As a decoder, image-removed contrast is exactly PMI re-ranking. It is at least as accurate as greedy decoding on POPE on all three models, with significant gains on four of nine model--split pairs; on LLaVA-1.5-7B a zero-threshold gate retains the gain with far fewer edits, and the correction does not reduce hallucination in long captions. As a source of training negatives, the blind stream lowered POPE accuracy on LLaVA-1.5-7B, did not significantly change CHAIR, and its BTS weighting had no measurable effect.

\FloatBarrier
\section*{Acknowledgments}
This work was carried out by a single author without institutional funding; compute was self-funded on Google Colab. The author used Claude (Anthropic), an AI assistant, for code development, statistical analysis, literature search and manuscript drafting. The author reviewed the results and the manuscript and takes responsibility for its content.

\section*{Impact Statement}
This work measures how much vision-language models depend on their visual input and evaluates corrections based on that measurement. The corrections studied had small effects on closed-form answers and did not reduce hallucination in captions, so they should not be relied on to make VLM output accurate. The work introduces no new model capability.

\section*{Reproducibility and Data Availability}
Code, notebooks and the per-sample logs from which the tables and figures are computed are available at \url{https://github.com/skytinstone/BTS-VLM}. All datasets used are public (VQAv2, POPE/COCO). The author declares no conflict of interest and no external funding.

\bibliographystyle{plainnat}
\bibliography{references}

\begin{thebibliography}{39}
\providecommand{\natexlab}[1]{#1}
\providecommand{\url}[1]{\texttt{#1}}
\expandafter\ifx\csname urlstyle\endcsname\relax
  \providecommand{\doi}[1]{doi: #1}\else
  \providecommand{\doi}{doi: \begingroup \urlstyle{rm}\Url}\fi

\bibitem[Agrawal et~al.(2018)Agrawal, Batra, Parikh, and
  Kembhavi]{agrawal2018dont}
Aishwarya Agrawal, Dhruv Batra, Devi Parikh, and Aniruddha Kembhavi.
\newblock Don't just assume; look and answer: Overcoming priors for visual
  question answering.
\newblock In \emph{CVPR}, 2018.
\newblock arXiv:1712.00377.

\bibitem[Bai et~al.(2024)]{bai2024survey}
Zechen Bai et~al.
\newblock Hallucination of multimodal large language models: A survey.
\newblock \emph{arXiv preprint arXiv:2404.18930}, 2024.

\bibitem[Chen et~al.(2026)Chen, Zhu, Zhu, Sheng, Huang, Ji, and
  Zhang]{chen2026sdpr}
Pengxu Chen, Yao Zhu, Guangming Zhu, Jun Sheng, Jincai Huang, Xiangyang Ji, and
  Liang Zhang.
\newblock Look clearly before answering: Mitigating hallucinations in {LVLMs}
  via saliency-driven perceptual realignment.
\newblock \emph{arXiv preprint arXiv:2607.16841}, 2026.

\bibitem[Chen et~al.(2025)Chen, Zhang, Liu, Wu, Zhang, and Tan]{chen2025mod}
Xinlong Chen, Yuanxing Zhang, Qiang Liu, Junfei Wu, Fuzheng Zhang, and Tieniu
  Tan.
\newblock Mixture of decoding: An attention-inspired adaptive decoding strategy
  to mitigate hallucinations in large vision-language models.
\newblock In \emph{Findings of ACL}, 2025.
\newblock arXiv:2505.17061.

\bibitem[Chen et~al.(2024)]{chen2024halc}
Zhaorun Chen et~al.
\newblock {HALC}: Object hallucination reduction via adaptive focal-contrast
  decoding.
\newblock In \emph{ICML}, 2024.
\newblock arXiv:2403.00425.

\bibitem[Fang et~al.(2025)Fang, Zhou, Kong, Gao, Chen, and Xia]{fang2025cpmi}
Hao Fang, Changle Zhou, Jiawei Kong, Kuofeng Gao, Bin Chen, and Shu-Tao Xia.
\newblock Grounding language with vision: A conditional mutual information
  calibrated decoding strategy for reducing hallucinations in {LVLMs}.
\newblock \emph{arXiv preprint arXiv:2505.19678}, 2025.

\bibitem[Favero et~al.(2024)Favero, Zancato, Trager, Choudhary, Perera,
  Achille, Swaminathan, and Soatto]{favero2024m3id}
Alessandro Favero, Luca Zancato, Matthew Trager, Siddharth Choudhary,
  Pramuditha Perera, Alessandro Achille, Ashwin Swaminathan, and Stefano
  Soatto.
\newblock Multi-modal hallucination control by visual information grounding.
\newblock In \emph{CVPR}, 2024.
\newblock arXiv:2403.14003.

\bibitem[Goyal et~al.(2017)Goyal, Khot, Summers-Stay, Batra, and
  Parikh]{goyal2017vqav2}
Yash Goyal, Tejas Khot, Douglas Summers-Stay, Dhruv Batra, and Devi Parikh.
\newblock Making the {V} in {VQA} matter: Elevating the role of image
  understanding in visual question answering.
\newblock In \emph{CVPR}, 2017.
\newblock arXiv:1612.00837.

\bibitem[Guan et~al.(2024)]{guan2024hallusionbench}
Tianrui Guan et~al.
\newblock {HallusionBench}: An advanced diagnostic suite for entangled language
  hallucination and visual illusion in large vision-language models.
\newblock In \emph{CVPR}, 2024.
\newblock arXiv:2310.14566.

\bibitem[Ho and Salimans(2022)]{ho2022cfg}
Jonathan Ho and Tim Salimans.
\newblock Classifier-free diffusion guidance.
\newblock \emph{arXiv preprint arXiv:2207.12598}, 2022.

\bibitem[Hu et~al.(2022)Hu, Shen, Wallis, Allen-Zhu, Li, Wang, Wang, and
  Chen]{hu2022lora}
Edward~J. Hu, Yelong Shen, Phillip Wallis, Zeyuan Allen-Zhu, Yuanzhi Li, Shean
  Wang, Lu~Wang, and Weizhu Chen.
\newblock {LoRA}: Low-rank adaptation of large language models.
\newblock In \emph{ICLR}, 2022.
\newblock arXiv:2106.09685.

\bibitem[Huang et~al.(2024)]{huang2024opera}
Qidong Huang et~al.
\newblock {OPERA}: Alleviating hallucination in multi-modal large language
  models via over-trust penalty and retrospection-allocation.
\newblock In \emph{CVPR}, 2024.
\newblock arXiv:2311.17911.

\bibitem[Huang et~al.(2026)Huang, Zhang, and Luo]{huang2026chasd}
Xiaoyi Huang, Kejia Zhang, and Zhiming Luo.
\newblock {CHASD}: Language increment-calibrated contrastive decoding against
  hallucination in {LVLMs}.
\newblock \emph{arXiv preprint arXiv:2605.23344}, 2026.

\bibitem[Lee et~al.(2024)Lee, Park, Jo, and Seo]{lee2024volcano}
Seongyun Lee, Sue~Hyun Park, Yongrae Jo, and Minjoon Seo.
\newblock Volcano: Mitigating multimodal hallucination through self-feedback
  guided revision.
\newblock In \emph{NAACL}, 2024.
\newblock arXiv:2311.07362.

\bibitem[Leng et~al.(2024)Leng, Zhang, Chen, Li, Lu, Miao, and
  Bing]{leng2024vcd}
Sicong Leng, Hang Zhang, Guanzheng Chen, Xin Li, Shijian Lu, Chunyan Miao, and
  Lidong Bing.
\newblock Mitigating object hallucinations in large vision-language models
  through visual contrastive decoding.
\newblock In \emph{CVPR}, 2024.
\newblock arXiv:2311.16922.

\bibitem[Li et~al.(2023{\natexlab{a}})]{li2023contrastive}
Xiang~Lisa Li et~al.
\newblock Contrastive decoding: Open-ended text generation as optimization.
\newblock In \emph{ACL}, 2023{\natexlab{a}}.
\newblock arXiv:2210.15097.

\bibitem[Li et~al.(2023{\natexlab{b}})Li, Du, Zhou, Wang, Zhao, and
  Wen]{li2023pope}
Yifan Li, Yifan Du, Kun Zhou, Jinpeng Wang, Wayne~Xin Zhao, and Ji-Rong Wen.
\newblock Evaluating object hallucination in large vision-language models.
\newblock In \emph{EMNLP}, 2023{\natexlab{b}}.
\newblock arXiv:2305.10355.

\bibitem[Lin(1991)]{lin1991divergence}
Jianhua Lin.
\newblock Divergence measures based on the {S}hannon entropy.
\newblock \emph{IEEE Transactions on Information Theory}, 37\penalty0
  (1):\penalty0 145--151, 1991.

\bibitem[Liu et~al.(2024{\natexlab{a}})Liu, Lin, Li, Wang, Yacoob, and
  Wang]{liu2024lrv}
Fuxiao Liu, Kevin Lin, Linjie Li, Jianfeng Wang, Yaser Yacoob, and Lijuan Wang.
\newblock Mitigating hallucination in large multi-modal models via robust
  instruction tuning.
\newblock In \emph{ICLR}, 2024{\natexlab{a}}.
\newblock arXiv:2306.14565.

\bibitem[Liu et~al.(2023)Liu, Li, Wu, and Lee]{liu2023llava}
Haotian Liu, Chunyuan Li, Qingyang Wu, and Yong~Jae Lee.
\newblock Visual instruction tuning.
\newblock In \emph{NeurIPS}, 2023.
\newblock arXiv:2304.08485.

\bibitem[Liu et~al.(2024{\natexlab{b}})Liu, Li, Li, and Lee]{liu2024llava15}
Haotian Liu, Chunyuan Li, Yuheng Li, and Yong~Jae Lee.
\newblock Improved baselines with visual instruction tuning.
\newblock In \emph{CVPR}, 2024{\natexlab{b}}.
\newblock arXiv:2310.03744.

\bibitem[Liu et~al.(2024{\natexlab{c}})Liu, Zheng, and Chen]{liu2024pai}
Shi Liu, Kecheng Zheng, and Wei Chen.
\newblock Paying more attention to image: A training-free method for
  alleviating hallucination in {LVLMs}.
\newblock In \emph{ECCV}, 2024{\natexlab{c}}.
\newblock arXiv:2407.21771.

\bibitem[Liu et~al.(2026)Liu, Wen, Lan, Wang, and Su]{liu2026selfval}
Shiyu Liu, Xinyi Wen, Zhibin Lan, Ante Wang, and Jinsong Su.
\newblock Countering the over-reliance trap: Mitigating object hallucination
  for {LVLMs} via a self-validation framework.
\newblock \emph{arXiv preprint arXiv:2601.22451}, 2026.

\bibitem[Rafailov et~al.(2023)Rafailov, Sharma, Mitchell, Ermon, Manning, and
  Finn]{rafailov2023dpo}
Rafael Rafailov, Archit Sharma, Eric Mitchell, Stefano Ermon, Christopher~D.
  Manning, and Chelsea Finn.
\newblock Direct preference optimization: Your language model is secretly a
  reward model.
\newblock In \emph{NeurIPS}, 2023.
\newblock arXiv:2305.18290.

\bibitem[Rohrbach et~al.(2018)Rohrbach, Hendricks, Burns, Darrell, and
  Saenko]{rohrbach2018chair}
Anna Rohrbach, Lisa~Anne Hendricks, Kaylee Burns, Trevor Darrell, and Kate
  Saenko.
\newblock Object hallucination in image captioning.
\newblock In \emph{EMNLP}, 2018.
\newblock arXiv:1809.02156.

\bibitem[Sun et~al.(2024)]{sun2023llavarlhf}
Zhiqing Sun et~al.
\newblock Aligning large multimodal models with factually augmented {RLHF}.
\newblock In \emph{Findings of ACL}, 2024.
\newblock arXiv:2309.14525.

\bibitem[Suo et~al.(2025)Suo, Zhang, Sun, Wu, Wang, and Zhang]{suo2025octopus}
Wei Suo, Lijun Zhang, Mengyang Sun, Lin~Yuanbo Wu, Peng Wang, and Yanning
  Zhang.
\newblock Octopus: Alleviating hallucination via dynamic contrastive decoding.
\newblock In \emph{CVPR}, 2025.
\newblock arXiv:2503.00361.

\bibitem[Wang et~al.(2023)]{wang2023amber}
Junyang Wang et~al.
\newblock {AMBER}: An {LLM}-free multi-dimensional benchmark for {MLLMs}
  hallucination evaluation.
\newblock \emph{arXiv preprint arXiv:2311.07397}, 2023.

\bibitem[Wang et~al.(2024)Wang, Pan, Ding, and Biemann]{wang2024icd}
Xintong Wang, Jingheng Pan, Liang Ding, and Chris Biemann.
\newblock Mitigating hallucinations in large vision-language models with
  instruction contrastive decoding.
\newblock In \emph{Findings of ACL}, 2024.
\newblock arXiv:2403.18715.

\bibitem[Xie et~al.(2024)Xie, Li, Xu, and Kan]{xie2024vdpo}
Yuxi Xie, Guanzhen Li, Xiao Xu, and Min-Yen Kan.
\newblock {V-DPO}: Mitigating hallucination in large vision language models via
  vision-guided direct preference optimization.
\newblock In \emph{Findings of EMNLP}, 2024.
\newblock arXiv:2411.02712.

\bibitem[Ye et~al.(2026)Ye, Li, Feng, Chen, Li, Ren, Chen, Tu, and
  Qin]{ye2026pgpo}
Zekai Ye, Qiming Li, Xiaocheng Feng, Ruihan Chen, Ziming Li, Haoyu Ren, Kun
  Chen, Dandan Tu, and Bing Qin.
\newblock Not all tokens see equally: Perception-grounded policy optimization
  for large vision-language models.
\newblock \emph{arXiv preprint arXiv:2604.01840}, 2026.

\bibitem[Yin et~al.(2025)Yin, Si, and Wang]{yin2025mirage}
Hao Yin, Guangzong Si, and Zilei Wang.
\newblock The mirage of performance gains: Why contrastive decoding fails to
  mitigate object hallucinations in {MLLMs}?
\newblock \emph{arXiv preprint arXiv:2504.10020}, 2025.

\bibitem[Yin et~al.(2024)]{yin2023woodpecker}
Shukang Yin et~al.
\newblock Woodpecker: Hallucination correction for multimodal large language
  models.
\newblock \emph{Science China Information Sciences}, 2024.
\newblock arXiv:2310.16045.

\bibitem[Yu et~al.(2024)]{yu2024rlhfv}
Tianyu Yu et~al.
\newblock {RLHF-V}: Towards trustworthy {MLLMs} via behavior alignment from
  fine-grained correctional human feedback.
\newblock In \emph{CVPR}, 2024.
\newblock arXiv:2312.00849.

\bibitem[Zhao et~al.(2025)Zhao, Zhang, Sun, Kong, Tan, and Feng]{zhao2025cicd}
Jianfei Zhao, Feng Zhang, Xin Sun, Lingxing Kong, Zhixing Tan, and Chong Feng.
\newblock Cross-image contrastive decoding: Precise, lossless suppression of
  language priors in large vision-language models.
\newblock \emph{arXiv preprint arXiv:2505.10634}, 2025.

\bibitem[Zhao et~al.(2026)Zhao, Wang, Liu, and Tang]{zhao2026safe}
Zhipeng Zhao, Wenxu Wang, Peishun Liu, and Ruichun Tang.
\newblock What do hallucinations reveal about multimodal reasoning? diagnosing
  visual grounding failures via contrastive decoding probes.
\newblock \emph{arXiv preprint arXiv:2609.16646}, 2026.

\bibitem[Zhao et~al.(2023)]{zhao2023hadpo}
Zhiyuan Zhao et~al.
\newblock Beyond hallucinations: Enhancing {LVLMs} through hallucination-aware
  direct preference optimization.
\newblock \emph{arXiv preprint arXiv:2311.16839}, 2023.

\bibitem[Zhou et~al.(2024{\natexlab{a}})Zhou, Cui, Rafailov, Finn, and
  Yao]{zhou2024povid}
Yiyang Zhou, Chenhang Cui, Rafael Rafailov, Chelsea Finn, and Huaxiu Yao.
\newblock Aligning modalities in vision large language models via preference
  fine-tuning.
\newblock \emph{arXiv preprint arXiv:2402.11411}, 2024{\natexlab{a}}.

\bibitem[Zhou et~al.(2024{\natexlab{b}})]{zhou2024lure}
Yiyang Zhou et~al.
\newblock Analyzing and mitigating object hallucination in large
  vision-language models.
\newblock In \emph{ICLR}, 2024{\natexlab{b}}.
\newblock arXiv:2310.00754.

\end{thebibliography}

\clearpage
\ifonecol\else\onecolumn\fi
\appendix

\section{Proof of Proposition~\ref{prop:bts}}
\label{app:proof}
For any two distributions, $0\le\JSD(p\|q)\le\ln 2$, with equality on the left iff $p=q$ and on the right iff $p$ and $q$ have disjoint supports~\citep{lin1991divergence}. Hence $a_t := 1-\JSD(\pv\|\pb)/\ln 2\in[0,1]$, so $0\le\BTS_t=\pb(\hat y_t)\,a_t\le\pb(\hat y_t)\le 1$. If $\pv=\pb$ then $a_t=1$ and $\BTS_t=\pb(\hat y_t)$; conversely, if $\pb(\hat y_t)>0$ and $\BTS_t=\pb(\hat y_t)$ then $a_t=1$, i.e.\ $\pv=\pb$. If the supports are disjoint then $a_t=0$; if $\pb(\hat y_t)=0$ the product is zero. \hfill$\square$

\section{Proofs of Propositions~\ref{prop:flip} and~\ref{prop:binary}}
\label{app:proof2}
\paragraph{Proposition~\ref{prop:flip}.} Let $\lv_t(v)=\log\pv(v)+c_1$ and $\lb_t(v)=\log\pb(v)+c_2$. Then $\tilde\ell_t(v)=(1+\alpha)\log\pv(v)-\alpha\log\pb(v)+c=\log\pv(v)+\alpha\,\mathrm{PMI}_t(v)+c$, with $c$ independent of $v$. Since $\gamma\le 1$, $\hat y_t\in\mathcal{V}_t$. The output differs from $\hat y_t$ iff some $v\in\mathcal{V}_t\setminus\{\hat y_t\}$ has a strictly larger score, i.e.\ $\alpha(\mathrm{PMI}_t(v)-\mathrm{PMI}_t(\hat y_t))>\log\pv(\hat y_t)-\log\pv(v)=\Delta_t(v)$. If $\pv=\pb$, all PMI values are zero and the left side is $0$, which cannot exceed $\Delta_t(v)\ge 0$. \hfill$\square$

\paragraph{Proposition~\ref{prop:binary}.} Suppose the correction changes $\hat y_t$ to $v$ and $\mathrm{PMI}_t(\hat y_t)\ge 0$. By Proposition~\ref{prop:flip}, $\mathrm{PMI}_t(v)>\mathrm{PMI}_t(\hat y_t)+\Delta_t(v)/\alpha\ge 0$, so $\pv(v)>\pb(v)$. Since $\pv(\hat y_t)+\pv(v)\le 1=\pb(\hat y_t)+\pb(v)+R_t(v)$, Eq.~\eqref{eq:residual} follows. If $R_t(v)=0$, the left side of Eq.~\eqref{eq:residual} is strictly positive because its first bracket is non-negative and its second is positive, which is a contradiction; hence $\mathrm{PMI}_t(\hat y_t)<0$. When also $\hat y_t\in\mathcal{C}$, this is the condition under which the PMI gate with $\tau=0$ is open. \hfill$\square$

\section{Experimental details}
\label{app:details}

\begin{table}[h]
\caption{Notation.}
\label{tab:notation}
\centering\small
\begin{tabular}{@{}ll@{}}
\toprule
Symbol & Meaning \\
\midrule
$I$, $x$, $y$ & image, text prompt, generated answer \\
$\pv$, $\pb$ & next-token distribution with / without the image \\
$\lv_t$, $\lb_t$ & corresponding logits at step $t$ \\
$\hat y_t$ & visual top-1 token, $\arg\max_v\pv(v)$ \\
$\BTS_t$ & Blind Trust Syndrome score, Eq.~\eqref{eq:bts} \\
$y^{+}$, $y^{-}$ & reference answer, blind (dispreferred) answer \\
$\ptheta$, $\pref$ & trained policy, frozen reference \\
$\beta$, $\lambda$ & DPO temperature, BTS weight strength \\
$\tau$, $\alpha$, $\gamma$ & gate threshold, contrast strength, plausibility cut-off \\
\bottomrule
\end{tabular}
\end{table}

\begin{table}[h]
\caption{Hyper-parameters.}
\label{tab:hparams}
\centering\small
\begin{tabular}{@{}ll@{}}
\toprule
Item & Value \\
\midrule
Models & LLaVA-1.5-7B/13B, LLaVA-NeXT-Mistral-7B, bf16 \\
LoRA & $r=16$, $\alpha=32$, dropout 0.05, q/k/v/o of the language model \\
Optimizer & AdamW, lr $5\times10^{-5}$, grad.\ accumulation 8, 1 epoch \\
PAPO & $\beta=0.1$, $\lambda\in\{1,0\}$, 3{,}000 VQAv2 questions (2{,}347 pairs) \\
TGSC (7B) & PMI gate, $\tau=0$, $\alpha=3$, $\gamma=0.1$ \\
TGSC (13B, NeXT) & PMI gate, $\tau=0$, $\alpha=1$, $\gamma=0.1$ \\
POPE (7B, 13B) & 2{,}400 test and 600 validation questions per split \\
POPE (NeXT) & 1{,}000 test and 300 validation questions per split \\
CHAIR & 500 COCO val2014 images, $\le 512$ new tokens \\
Diagnostic captions & 50 COCO images, $\le 96$ new tokens \\
\bottomrule
\end{tabular}
\end{table}

\paragraph{Compute.} One A100 ran the full suite in about 8 hours: LLaVA-1.5-7B 4.8\,h (including PAPO for both values of $\lambda$), LLaVA-1.5-13B 1.6\,h and NeXT 1.7\,h, excluding model download and a preliminary short test.

\clearpage
\section{Additional results}
\label{app:results}

\begin{table}[!htbp]
\caption{Gate selection on 600 POPE-adversarial validation questions (LLaVA-1.5-7B, $\alpha=1$, $\gamma=0.1$). Help / harm: questions fixed / broken relative to greedy decoding.}
\label{tab:tau}
\centering\small
\begin{tabular}{@{}lcccccccc@{}}
\toprule
Setting & $\tau$ & Acc. & F1 & FP & FN & Help & Harm & Gate rate (\%) \\
\midrule
Greedy & -- & 81.83 & 81.24 & 35 & 74 & 0 & 0 & 0.0 \\
Always-on & -- & 82.00 & 81.38 & 34 & 74 & 2 & 1 & 100.0 \\
BTS gate & 0.20 & 82.00 & 81.38 & 34 & 74 & 2 & 1 & 44.1 \\
BTS gate & 0.35 & 82.00 & 81.38 & 34 & 74 & 2 & 1 & 33.7 \\
BTS gate & 0.50 & 82.00 & 81.38 & 34 & 74 & 2 & 1 & 10.6 \\
BTS gate & 0.65 & 81.67 & 81.10 & 36 & 74 & 0 & 1 & 4.1 \\
BTS gate & 0.80 & 81.83 & 81.24 & 35 & 74 & 0 & 0 & 3.3 \\
PMI gate & 0.00 & 82.00 & 81.38 & 34 & 74 & 2 & 1 & 3.4 \\
PMI gate & 0.10 & 81.83 & 81.24 & 35 & 74 & 1 & 1 & 0.7 \\
PMI gate & 0.25 & 81.67 & 81.10 & 36 & 74 & 0 & 1 & 0.2 \\
PMI gate & 0.50 & 81.83 & 81.24 & 35 & 74 & 0 & 0 & 0.1 \\
\bottomrule
\end{tabular}
\end{table}

\begin{table}[!h]
\caption{Examples of PAPO preference pairs (dispreferred $=$ blind response).}
\label{tab:pairs}
\centering\small
\begin{tabular}{@{}lll@{}}
\toprule
Question & Preferred & Dispreferred \\
\midrule
What are the people in the background doing? & watching & Standing \\
What is he on top of? & picnic table & Roof \\
What website copyrighted the picture? & foodiebakercom & Unknown \\
What is to the right of the soup? & chopsticks & Bread \\
What does the truck on the left sell? & ice cream & Food \\
Why is the cow laying down? & tired & Resting \\
How many are playing ball? & 1 & 2 \\
How many pictures are there? & 7 & 1 \\
\bottomrule
\end{tabular}
\end{table}

\begin{figure}[!htbp]
\centering
\includegraphics[width=\textwidth]{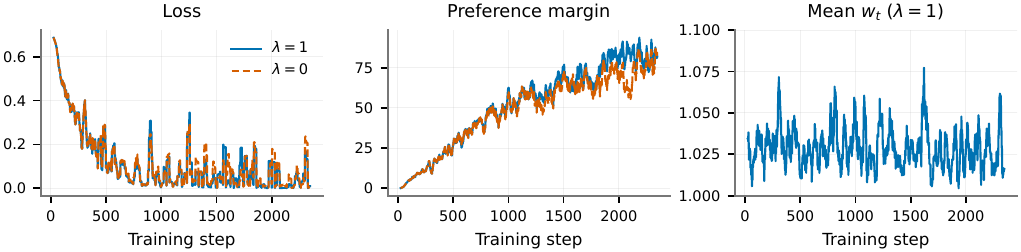}
\caption{PAPO training on LLaVA-1.5-7B; 25-step moving average.}
\label{fig:train}
\end{figure}

\begin{figure}[!htbp]
\centering
\includegraphics[width=\textwidth]{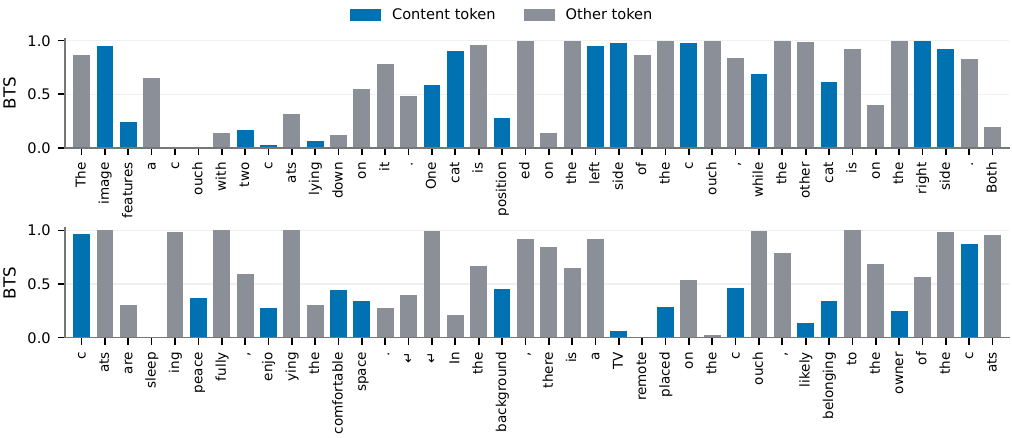}
\caption{Per-token BTS score on a COCO caption generated by LLaVA-1.5-7B.}
\label{fig:token}
\end{figure}

\begin{figure}[!htbp]
\centering
\includegraphics[width=0.8\textwidth]{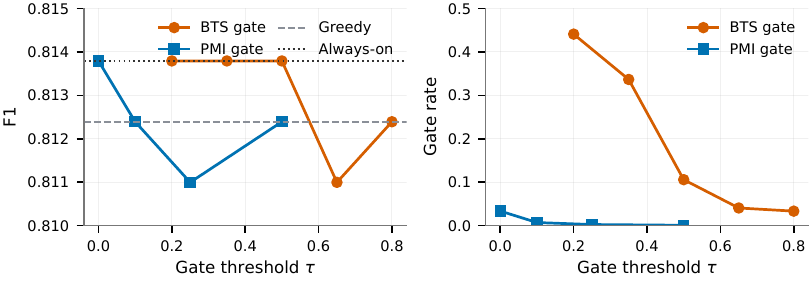}
\caption{Gate threshold sweep on the 600 POPE-adversarial validation questions. Differences in F1 amount to one to three questions, and its axis is strongly magnified.}
\label{fig:tau}
\end{figure}

\end{document}